\documentclass[10pt,twocolumn,letterpaper]{article}

\usepackage{iccv}
\usepackage{times}
\usepackage{epsfig}
\usepackage{graphicx}
\usepackage{amsmath}
\usepackage{amssymb}

\usepackage{multirow}
\usepackage{textcomp} 
\usepackage{mathtools}
\usepackage{multirow}
\usepackage{booktabs}
\usepackage[pagebackref=true,breaklinks=true,letterpaper=true,colorlinks,bookmarks=false]{hyperref}

\iccvfinalcopy 

\def\iccvPaperID{2510} 

\ificcvfinal\fi
\begin{document}

\title{What and Where: Modeling Skeletons from Semantic and Spatial Perspectives for Action Recognition}

\author{
	Lei Shi$^{1,2}$ \and Yifan Zhang$^{1,2}$\thanks{Corresponding Author} \and Jian Cheng$^{1,2,3}$ \and Hanqing Lu$^{1,2}$ \and
	$^1$National Laboratory of Pattern Recognition, Institute of Automation, Chinese Academy of Sciences\\
	$^2$University of Chinese Academy of Sciences\\
	$^3$CAS Center for Excellence in Brain Science and Intelligence Technology\\
	{\tt\small \{lei.shi, yfzhang, jcheng, luhq\}@nlpr.ia.ac.cn} 
}
\maketitle
\ificcvfinal\thispagestyle{empty}\fi

\begin{abstract}
Skeleton data, which consists of only the 2D/3D coordinates of the human joints, has been widely studied for human action recognition. 
Existing methods take the semantics as prior knowledge to group human joints and draw correlations according to their spatial locations, which we call the semantic perspective for skeleton modeling.  
In this paper, in contrast to previous approaches, we propose to model skeletons from a novel spatial perspective, from which the model takes the spatial location as prior knowledge to group human joints and mines the discriminative patterns of local areas in a hierarchical manner.
The two perspectives are orthogonal and complementary to each other; and by fusing them in a unified framework, our method achieves a more comprehensive understanding of the skeleton data. 
Besides, we customized two networks for the two perspectives. 
From the semantic perspective, we propose a Transformer-like network that is expert in modeling joint correlations, and present three effective techniques to adapt it for skeleton data. 
From the spatial perspective, we transform the skeleton data into the sparse format for efficient feature extraction and present two types of sparse convolutional networks for sparse skeleton modeling. 
Extensive experiments are conducted on three challenging datasets for skeleton-based human action/gesture recognition, namely, NTU-60, NTU-120 and SHREC, where our method achieves state-of-the-art performance. 
\end{abstract}

\section{Introduction}

\begin{figure}[htp]
    \centering
    \includegraphics[width=1\linewidth]{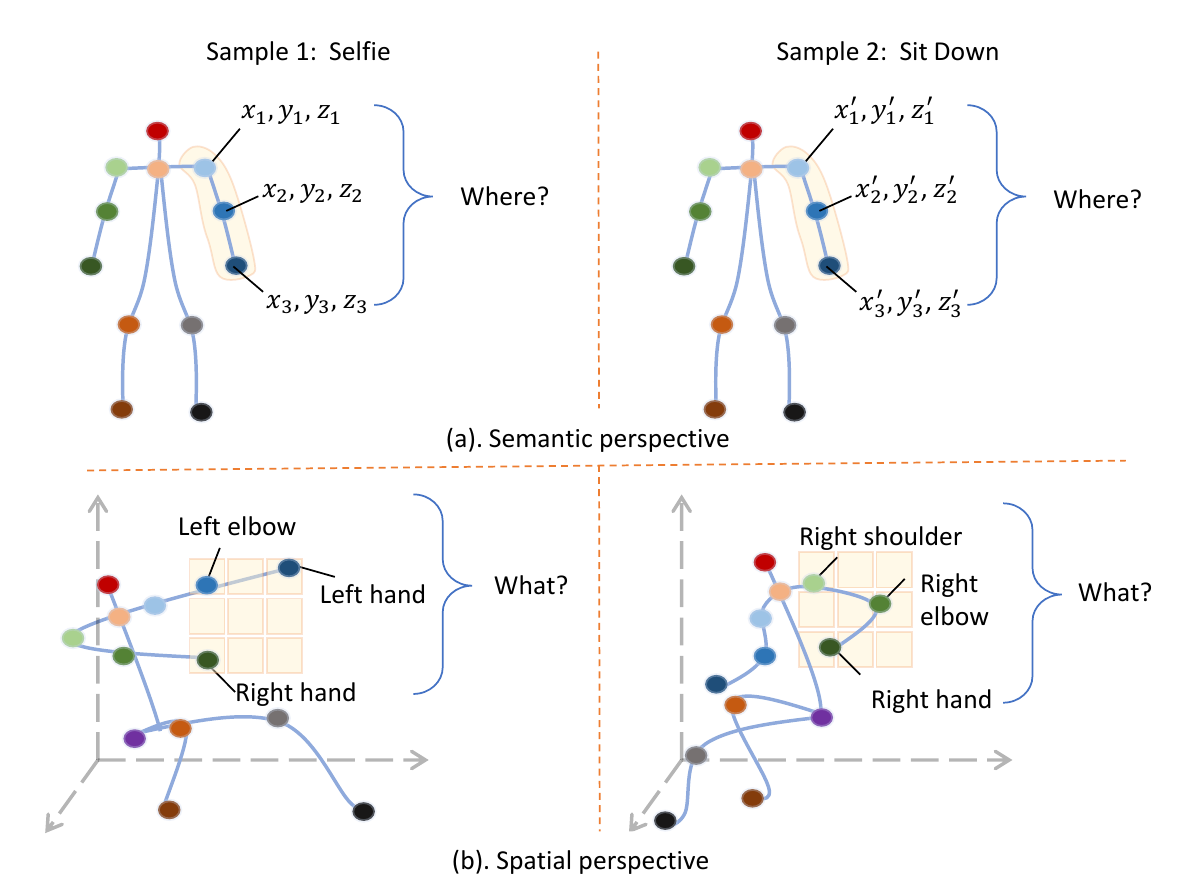}
    \caption{In a specific computing step, (a) 
    the SEM-Net only groups the left hand, the left elbow and the left shoulder to model their correlations, regardless of whether the sample category is ``selfie" or ``sit down"; 
    (b) 
    the SPA-Net mines discriminative patterns of the joints in the local areas, where these joints may be different for different samples. 
    }
    \label{fig:twoview}
\end{figure}{}

Action recognition is a popular research topic because it can be applied to many practical fields such as human-computer interaction and video surveillance~\cite{carreira_quo_2017,shi_gesture_2019,feichtenhofer_slowfast_2019,shi_action_2019}.
In recent years, skeleton-based action recognition has drawn considerable attention due to its small amount of data, higher-level semantic information and strong robustness for complicated environment~\cite{yan_spatial_2018,cho_self-attention_2020,shi_two-stream_2019,liu_disentangling_2020}. 
In contrast to RGB representations, skeleton data contains only the 2D or 3D locations of the human key joints, providing high-abstract and environment-free information, allowing action recognition algorithms to focus on the robust features of the action.

Recently, data-driven methods have become the mainstream for skeleton-based action recognition~\cite{yan_spatial_2018,shi_two-stream_2019,shi_skeleton-based_2019}, where the raw skeletons are transformed into different formats (e.g., pseudo-images and graphs) to be fed into different models (e.g., CNNs and GCNs). 
The transformation rules are usually designed based on the semantic categories of the human joints while the characteristics of the joints are represented by their 2D/3D coordinates. 
It means the model takes semantics as prior knowledge to group human joints and determine the computing flows of the input joints in the model. 
For example, in the famous ST-GCN~\cite{yan_spatial_2018}, because the left hand and the right hand are not adjacent in the human body, the ``left hand" joint and the ``right hand" joint are not connected in the constructed graph for all of the samples, even if in some cases the are very close in space. 
We call this the semantic perspective for skeleton modeling. 

In this paper, in contrast to previous methods, we propose a novel spatial perspective for skeleton modeling.
Fig.~\ref{fig:twoview} shows the differences between the two perspectives.
From the spatial perspective, we organize the input joints according to their spatial positions and instantiate each joint with its semantic label. 
That is, the model will tour every local areas in the coordinate space and 
learn to activate discriminative patterns inside every local receptive fields. 
In fact, the spatial perspective is used more frequently in other computer vision tasks.
For example, when using CNNs for image/video classification, the receptive field of the convolutional kernel is determined by the proximity of pixels instead of their RGB values; that is, the space information determines the computing flows of the pixels. 
Note that the proposed spatial perspective is orthogonal and complementary to the semantic perspective.
By fusing them in a two-stream framework, it achieves higher performance. 


Besides, we customized two networks (SEM-Net and SPA-Net) for two perspectives. 
(1) From the semantic perspective, the key is to model the spatial-temporal correlations between different human joints.
In the field of NLP, the Transformer model, which is built based on the self-attention mechanism, has been shown great superiority for relation modeling~\cite{vaswani_attention_2017,dai_transformer-xl_2019}. 
Thus, we propose to design a Transformer-like network to better model the correlations between the joints of the skeleton data. 
However, the skeleton data has two dimensions: the spatial dimension and the temporal dimension. Directly flattening it brings a lot of computational cost when computing the self-attention module, but processing the space and time separately cannot well model the inter-frame dependencies. 
In this paper, we propose a cross-frame self-attention layer (CFSA), which comes to a good compromise. 
Besides, different with the character strings in NLP, each joint of the skeletons has a specific semantic meaning (e.g., the hand, the neck, etc.), but directly use the vanilla transformer cannot well exploit this prior knowledge. 
In this work, we introduce a global regularization (GR) to regularize the correlations between specific pairs of semantics, which brings notable improvement. 
(2) From the spatial perspective, the key is to judge whether the joints inside a local receptive filed constitute a discriminative pattern for a certain category. 
CNN is a good choice for this task, which has been proved in many computer vision tasks~\cite{krizhevsky_imagenet_2012,cao_realtime_2017,long_fully_2015}. 
However, because the locations of the skeleton points in a 3D space are very sparse, directly performing dense convolution will cause the waste of the computation. 
In this work, inspired by \cite{choy_4d_2019}, we propose to transform the raw data into the sparse format and present two types of sparse convolutional networks for efficient sparse skeleton modeling. 
The first is the sparse 3+1D convolutional network (SPA-3+1D), which employs a sparse 3D convolution to model spatial features and append a 1D temporal convolution to capture temporal dependencies. 
The second is the sparse 4D convolutional network (SPA-4D), which can directly extract the spatial-temporal features for sparse skeleton joints from low-levels to high-levels. 

To verify the effectiveness of the proposed method, extensive experiments are conducted on NTU-60~\cite{shahroudy_ntu_2016}, NTU-120~\cite{liu_ntu_2019} and SHREC~\cite{de_smedt_shrec17_2017}, where our method achieves state-of-the-art performance on both of them. Code will be released.

Overall, our contributions lie in three aspects. 
\begin{enumerate}
    \item We propose a new way for looking at the problem of skeleton modeling, namely, the spatial perspective, which is orthogonal and complementary to the previous semantic perspective. 
    We make extensive experiments to verify the feasibility of the spatial perspective for skeleton data. 
    By fusing the two perspectives in a unified framework, our method achieves state-of-the-art performance on NTU-60, NTU-120 and SHREC. 
    \item From the semantic perspective, we propose a Transformer-like network that is expert in modeling joint correlations, and present effective techniques to adapt it for skeleton data.
    \item From the spatial perspective, we transform the skeleton data into the sparse format for efficient feature extraction and propose two types of sparse convolutional networks (SPA-4D and SPA-3+1D) for sparse skeleton modeling.
\end{enumerate}

\section{Related Works}
\subsection{Skeleton-Based Action Recognition}
Skeleton-based action recognition has been studied for decades. 
The mainstream approaches in recent years lie in three aspect: 
(1) the RNN-based approaches where the skeleton sequence is fed into the RNN models and the joints are modeled in a predefined traversal order~\cite{zhang_view_2017,li_independently_2018,si_skeleton-based_2018,si_attention_2019}. 
(2) the CNN-based approaches where the skeleton sequence is transformed into a pseudo-image and is fed into the CNNs for recognition~\cite{li_skeleton-based_2017,liu_enhanced_2017,cao_skeleton-based_2018}. 
(3) the GCN-based approaches where the skeleton sequence is encoded into a spatiotemporal graph according to the physical structure of the human body and is modeled with GCNs~\cite{yan_spatial_2018,tang_deep_2018,shi_two-stream_2019,shi_skeleton-based_2019}. 
Recently, GCN-based approaches have shown better performance compared with other methods. 
Yan et al.~\cite{yan_spatial_2018} are the first to use GCN for skeleton-based action recognition. 
Shi et al.~\cite{shi_two-stream_2019} propose an adaptive graph convolutional network to adapt the graph building process. They also introduce a two-stream framework to use both the joint information and the bone information.
Peng et al.~\cite{peng_learning_2020} turn to Neural Architecture Search (NAS) to automatically design the architecture of the network. 
Liu et al.~\cite{liu_disentangling_2020} disentangle the importance of nodes in different neighborhoods for effective long-range modeling. They also leverage the features of adjacent frames to capture complex spatial-temporal dependencies. 
Besides, some works are trying to speed up the GCN for skeleton data. 
For example, Cheng et al.~\cite{cheng_skeleton-based_2020} introduce a shift operator to replace the heavy regular graph convolutions. Zhang et al.~\cite{zhang_semantics-guided_2020} propose to exploit high-level information to reduce the depth of the network. 
However, none of these methods have try to view the skeletons from the spatial perspective; in contrast, we present a unified framework for modeling skeletons from both the semantic perspective and the spatial perspective, which achieves higher performance. 

\subsection{Self-Attention and Transformer}
Self-attention mechanism is the basic block of transformer~\cite{vaswani_attention_2017,dai_transformer-xl_2019}, which has shown great success in recent works of the NLP field. 
Its input consists of a set of queries $Q$, keys $K$ of dimension $C$ and values $V$, which are packaged in the matrix form for fast computation. 
It first computes the dot products of the query with all keys, divides each by $\sqrt{C}$, and applies a softmax function to obtain the weights on the values~\cite{vaswani_attention_2017}. 
In formulation: 
\begin{equation}
    Attention(Q, K, V) = softmax(\frac{QK^T}{\sqrt{C}})V
\end{equation}
The similar idea has also been used for many computer vision tasks such as relation modeling~\cite{santoro_simple_2017}, detection~\cite{hu_relation_2018} and semantic segmentation~\cite{fu_dual_2019}. 
For skeleton data, Cho et al.~\cite{cho_self-attention_2020} directly employ the transformer network for skeleton-based action recognition. 
However, Cho et al. only calculate the intra-frame correlations of the joints and neglect the inter-frame dependencies. 

\section{Method}
\begin{figure}[!htb]
    \centering
    \includegraphics[width=0.9\linewidth]{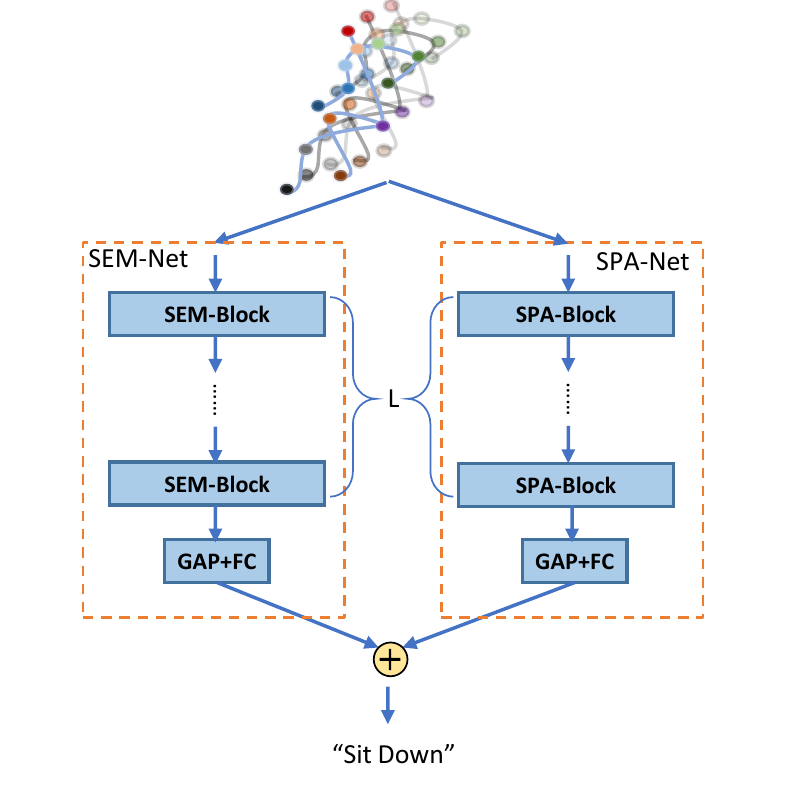}
    \caption{Overview of the pipeline. The input skeleton is modeled from both the semantics and spatial perspectives by SEM-Net and SPA-Net, respectively. GAP and FC represent the global-average-pooling layer and the fully-connected layer, respectively. }
    \label{fig:pipeline}
\end{figure}
Fig.~\ref{fig:pipeline} shows the overall architecture of the proposed method.
The input skeleton sequence is modeled from both the semantic perspective and the spatial perspective by SEM-Net and SPA-Net, respectively. 
Both of the two networks are stacked with $L$ basic blocks.
The softmax scores of two streams are averaged to get the final prediction label. 

\subsection{Semantic Perspective}
From the semantic perspective, the input is a sequence of joints each has a  specific meaning (e.g., hand or head). 
The key is to model the spatial-temporal correlations between different human joints based on their locations to inference the action. 
As introduced in related works, transformer (or self-attention module) is especially suitable for relation modeling. 
It mainly contains two modules: the self-attention layer and the feed-forward layer. 
To adapt it for the skeleton data, we replace them with the global-regularized cross-frame self-attention layer (GR-CFSA) and the multi-scale temporal feed-forward layer (MS-TFF), respectively.

\paragraph{Cross-Frame Self-Attention Layer}

\begin{figure}[!htb]
    \centering
    \includegraphics[width=0.9\linewidth]{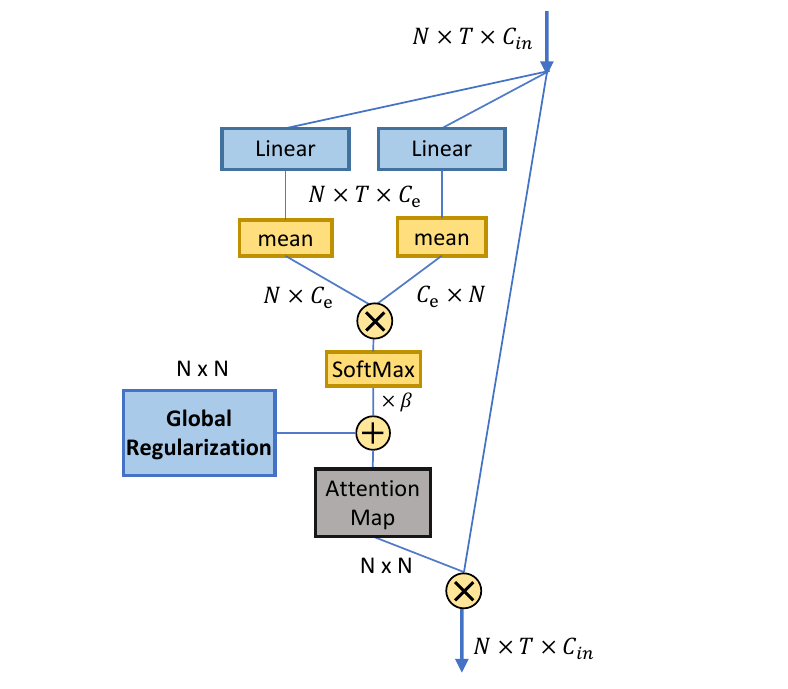}
    \caption{Illustration of the proposed global-regularized cross-frame self-attention layer (GR-CFSA). Blue box contains learnable parameters. }
    \label{fig:attention}
\end{figure}

The raw skeleton sequence can be represented as a joint set $\mathcal{X}=\{ \mathbf{x}_{t,n}\in\mathbb{R}^{C} | t,n \in\mathbb{Z}, 1\leq t\leq T, 1\leq n\leq N \}$. 
$T$ denotes the number of frames and $N$ denotes the number of joints. 

The raw skeleton data has two dimensions (D): the spatial dimension and the temporal dimension. 
To implement self-attention module for skeleton data, the most direct way is to flatten it into a 1-D sequence. 
In formulation: 
\begin{equation}
    \mathbf{{x}_{t_1,n_1}^{out}} = \sum_{t_2}^T\sum_{n_2}^N \sigma(\frac{\alpha_{t_1, n_1, t_2, n_2}}{\sqrt{C_e}})\mathbf{{x}_{t_2,n_2}^{in}}
    \label{eq:attentionU}
\end{equation}
where $\sigma$ denotes the SoftMax operation. $\alpha_{t_1, n_1, t_2, n_2}$ denotes the attention weight of the joint $\mathbf{x}_{t_1, n_1}$ and the joint $\mathbf{x}_{t_2, n_2}$, which is computed by:
\begin{equation}
    \alpha_{t_1, n_1, t_2, n_2} = f_1(\mathbf{{x}_{t_1,n_1}})f_2(\mathbf{{x}_{t_2,n_2}})^T
    \label{eq:weightU}
\end{equation}
where $f_1$ and $f_2$ are two embedding functions $f: C_{in}\longrightarrow C_{e}$. $C_{in}$ and $C_{e}$ denote input channels and embedding channels. 

However, the computational complexity of the self-attention module in this manner is $O(T^2N^2C_e)$, which is too expensive and makes the model difficult to optimize. 
Some works~\cite{cho_self-attention_2020} propose to reduce the computational cost by considering only the intra-frame correlations.
They reformulate the Eq.~\ref{eq:attentionU} to: 
\begin{equation}
    \mathbf{{x}_{t,n_1}^{out}} = \sum_{n_2}^N \sigma(\frac{\alpha_{t, n_1, n_2}}{\sqrt{C_e}})\mathbf{{x}_{t,n_2}^{in}}
    \label{eq:attention}
\end{equation}
where the attention weight $\alpha_{t, n_1, n_2}$ is computed by:
\begin{equation}
    \alpha_{t, n_1, n_2} = f_1(\mathbf{{x}_{t,n_1}})f_2(\mathbf{{x}_{t,n_2}})^T
    \label{eq:weight}
\end{equation}
In this way, the computational complexity is reduced to $O(N^2TC_e)$. 
However, only the intra-frame information is not enough for action recognition, especially for highly dynamic actions. 
For example, to classify the order-dependent actions such as ``sitting down" versus ``standing up", cross-frame information is important. 
In order to model the cross-frame correlations of the joints as well as keeping the computational cost smaller, we propose a compromise, where the Eq.~\ref{eq:weight} is reformulated to: 
\begin{equation}
\label{eq:newweight}   
\begin{aligned}
     \alpha_{t, n_1, n_2} 
    &= \sum_{t_1} \sum_{t_2} f_1(\mathbf{{x}_{t_1,n_1}})f_2(\mathbf{{x}_{t_2,n_2}})^T\\
    &= \sum_{\tau} f_1(\mathbf{{x}_{\tau,n_1}}) \sum_{\tau}f_2(\mathbf{{x}_{\tau,n_2}})^T
\end{aligned}
\end{equation}
In Eq.~\ref{eq:weight}, the correlations across every two frames are considered when calculating the attention weights, but through the simplification, the computing cost is not increased. 
Note that the Eq.~\ref{eq:attention} is not changed so that the features being aggregated is still limited to the same frame. 
Thus, the computational complexity is still $O(N^2TC_e)$. 

\paragraph{Global Regularization}
Different with the character strings that the conventional self-attention module usually modeled in the NLP tasks, 
each position of the skeleton data has a fixed human-body-based label, e.g., the head joint or the hand joint. 
It means there should be an inherent relationship between these positions. 
To represent this invariant relationship in the model, we propose a global regularization, 
which is an unified attention weight $\hat{\alpha}$ and is shared for all the frames and the data samples. 
It represents the inherent correlations between the corresponding two joints and is added to the original attention weight introduced in Eq.~\ref{eq:newweight}, which now becomes:
\begin{equation}
    \alpha_{t, n_1, n_2} = \beta\sum_{\tau} f_1(\mathbf{{x}_{\tau,n_1}}) \sum_{\tau}f_2(\mathbf{{x}_{\tau,n_2}})^T + \hat{\alpha_{n_1, n_2}}
    \label{eq:gloreg}
\end{equation}
where $\beta$ is a parameter used to balance the feature-based attentions and the inherent attentions. 

Fig.~\ref{fig:attention} shows the proposed global-regularized cross-frame self-attention layer (GR-CFSA) in the tensor form. 
The input data is formulated as a $N\times T\times C$ tensor.

\paragraph{Multi-Scale Feed-Forward Layer}
\begin{figure}[!htb]
    \centering
    \includegraphics[width=1\linewidth]{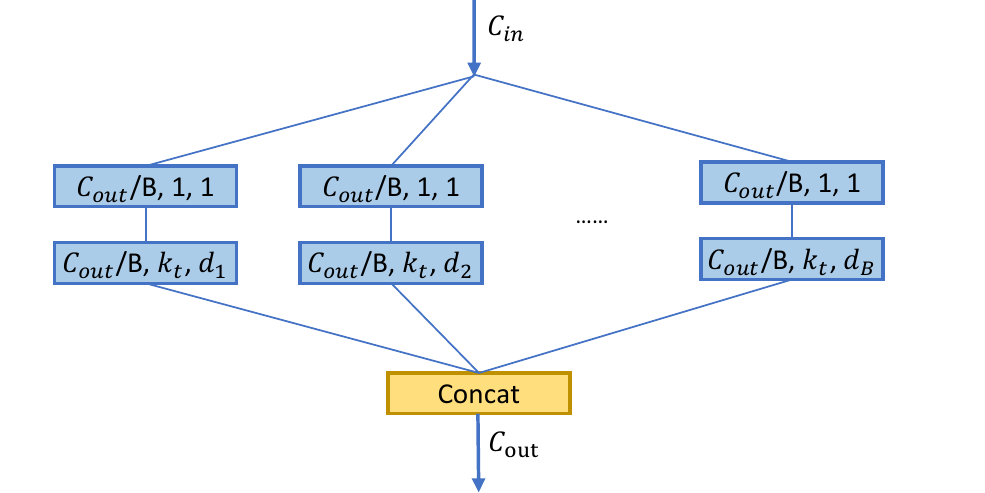}
    \caption{Illustration of the multi-scale temporal feed-forward layer (MS-TFF). $C_{out}$, $B$, $k$ and $d$ denote the output channels, branch number, kernel size and dilation value, respectively.  }
    \label{fig:tconv}
\end{figure}
Previous GCN-based works~\cite{yan_spatial_2018,shi_two-stream_2019} always add a temporal convolution after each graph convolutional layer, which have been verified very important for recognition. 
In this work, we replace the original feed-forward layer of the transformer with a temporal feed-forward (TFF) layer after every GR-CFSA layer. 
In detail, TFF is a convolutional layer where the kernel size along the spatial dimension and the temporal dimension is set to $1$ and $k_t$, respectively. 
We found that setting the size of the temporal window to a large value can cover more frames and improve the performance, but the computational cost also increases. 
In order to have larger temporal windows as well as saving the computational costs, we extend the vanilla TFF layer into a multi-scale TFF (MS-TFF) layer. 
Concretely, we split input channels into $B$ branches through a $1\times 1$ convolutional layer. 
Each branch has different temporal window size. 
Instead of using different $k_t$, we change the temporal window size by using different dilation values to reduce the amount of the parameters.
The output of all branches are concatenated to keep the output channels the same with the input. 
Fig.~\ref{fig:tconv} shows the architecture of the proposed MS-TFF layer.

\paragraph{SEM-Block}

\begin{figure}[!htb]
    \centering
    \includegraphics[width=0.9\linewidth]{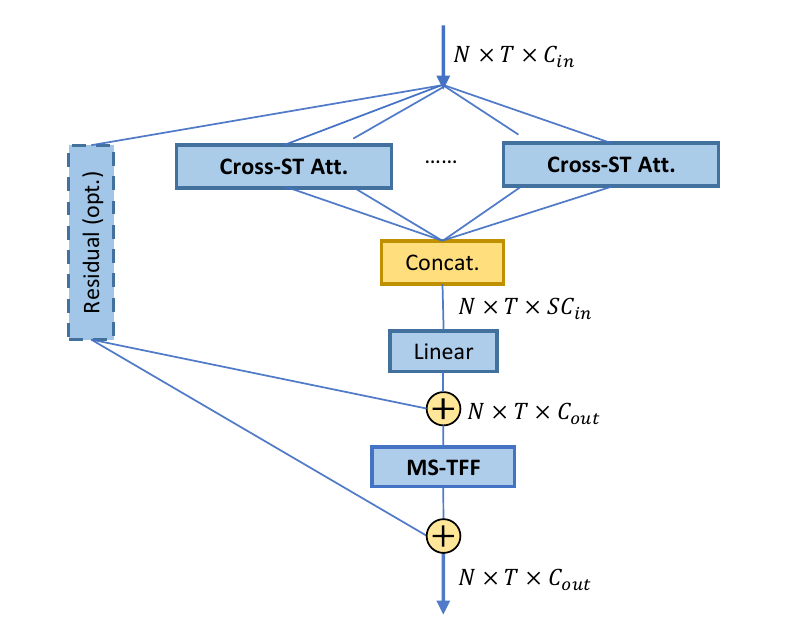}
    \caption{Illustration of the SEM-block. $S$ denote the number of the attention heads.}
    \label{fig:transformer}
\end{figure}

Fig.~\ref{fig:transformer} shows a single SEM-block, which is the basic block to build the final SEM-Net. 
Similar to the multi-head attention~\cite{vaswani_attention_2017}, it consists of $S$ parallel GR-CFSA layers, whose outputs are concatenated and aggregated through a linear layer. 
A MS-TFF layer is appended after the GR-CFSA layers. 
Residual connections are added in each block to ease the training. 
Finally, the SEM-Net is the stack of the SEM-Blocks. 


\paragraph{Bone stream}
For fair comparison with previous works~\cite{si_skeleton-based_2018,shi_two-stream_2019,shi_skeleton-based_2019}, we fuse the bone and the joint information to help improve the recognition performance. 
The bone information is a vector between two adjacent joints of the human body to represent the length and the direction of the human bones. 
The softmax scores of two streams are summed to obtain final prediction scores. 

\subsection{Spatial Perspective}

From the spatial perspective, the input is a local 3D space that contains a number of points. 
The key is to identify which joints these points are and whether they constitute a discriminative pattern for a certain category. 
To activate specific spatial patterns given a space, convolutional neural network is a good choice because it has been shown great performance in many computer vision tasks.
However, different with the pixels of images, the locations of the skeleton points in a 3D space are very sparse. 
Directly performing the convolution will cause the waste of the computation because most of the locations in the space is null. 
In this work, inspired by \cite{choy_4d_2019}, we propose to transform the skeleton joints into the sparse format and build a sparse convolutional network to model it efficiently. 

\paragraph{Sparse Convolution}
To perform sparse convolution, the raw skeleton data is first transformed into the sparse format as: 
\begin{equation}
    R = 
    \begin{bmatrix*}
    x_1& y_1& z_1& t_1 \\
    \vdots& \vdots& \vdots &\vdots  \\
    x_N& y_N& z_N& t_N \\
    \end{bmatrix*}, 
    F = 
    \begin{bmatrix*}
    \bf{f_1} \\
    \vdots \\
    \bf{f_N} \\
    \end{bmatrix*}
\end{equation}
where $R\in\mathbb{Z}^{N\times C_{coor}}$ and $F\in\mathbb{R}^{N\times C_{fea}}$ denote the coordinate matrix and the feature matrix, respectively. 
$C_{coor}=4$. 
$N$ is the number of the joints in a sample. 
x,y and z are the spatial coordinates of the points and t is the temporal index.
The raw skeleton data is a sequence of the joint coordinates. 
To transform it into the sparse format, the coordinates of the raw skeleton data are normalized into the range of zero to the space size and floored into integers, which constitute the coordinate matrix $R$. 
For feature matrix $F$, the feature of each point, i.e., $\mathbf{f_i}$, is represented as a one-hot vector that indicates which joint the current point is. 
For example, we can use $\mathbf{f}=[1, 0, 0]$ to represent the head joint and use $\mathbf{f}=[0, 1, 0]$ to represent the neck joint. 

Given $R$ and $F$, the sparse convolution is formulated as: 
\begin{equation}
    F_{out}(i) = \sum_{R(j)\in\mathcal{B}_{R(i)}} W_{R(i)-R(j)} F_{in}(j)
\end{equation}
where $R(i)$ denotes the coordinate of the $i_{th}$ point. $\mathcal{B}_{R(i)}$ denotes the neighbors of the point $i$ defined by the convolutional kernel. 
$W$ is the convoutional parameter. 
$W$ and $F_{in}$ are matched based on the relative position of the $j_{th}$ point to the $i_{th}$ point just like the conventional convolution. 
Similarly, sparse max-pooling operator is defined as: 
\begin{equation}
    F_{out}(i) = \max_{R(j)\in\mathcal{B}_{R(i)}} F_{in}(j)
\end{equation}

\paragraph{SPA-Block}
\begin{figure}[!htb]
    \centering
    \includegraphics[width=0.9\linewidth]{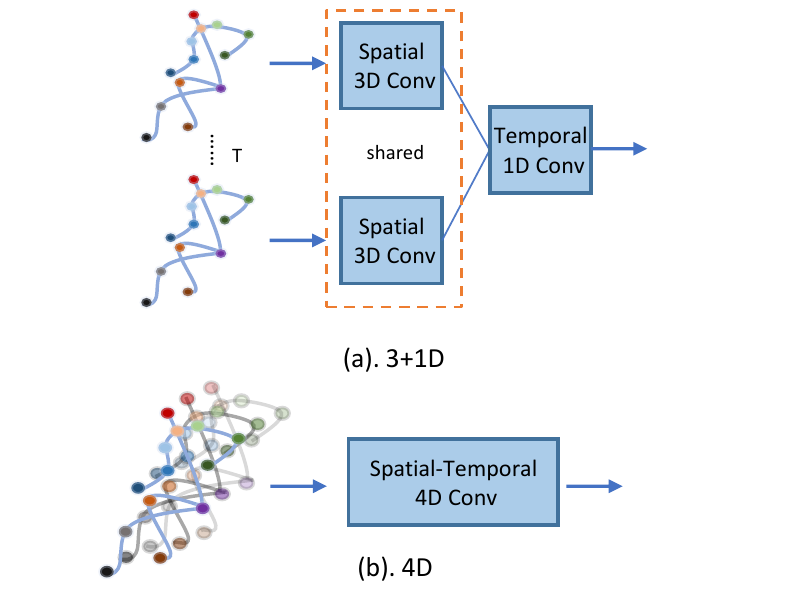}
    \caption{Illustration of the two strategies for SPA-block: (a) SPA-3+1D and (b) SPA-4D. }
    \label{fig:spa}
\end{figure}
With the definition of the sparse convolution and sparse max-pooling, we propose two strategies to build the basic SPA-Block as shown in Fig.~\ref{fig:spa}.  
The first is a 3+1D manner (SPA-3+1D), where each frame is modeled with a sparse 3D convolution to extract spatial features and then a temporal convolution is used to model temporal dependencies. 
The second is a 4D manner (SPA-4D), where the whole sequence is directly modeled with a sparse 4D convolution. 
The 4D convolution can directly model the spatial-temporal joint dependencies from low-levels to high-levels. 
However, it has more parameters due to the extra kernel dimension.
The relation between the 3+1D manner and the 4D manner is similar to the 3D convolutional networks versus 2+1D convolutional networks~\cite{qiu_learning_2017}. 

To strengthen the model's ability of modeling multi-scale temporal dependencies, we expend the temporal convolution to multi-scale temporal convolution. 
It is similar to the proposed MS-TFF layer in Fig.~\ref{fig:tconv}.
That is, we split the module into several branches, where each branch has different temporal window size controlled by different dilation values. 

Similarly, the SPA-Net is the stack of the SPA-Blocks. 


\section{Experiments}
We perform extensive experiments on three challenging datasets, namely, NTU-60, NTU-120 and SHREC. 
Detailed training schemes and network architectures are provided in the supplement material. 

\subsection{Datasets}
\textbf{NTU-60~\cite{shahroudy_ntu_2016}} consists of 56,578 action clips in 60 action classes. Each action is captured by 3 cameras. It provides 3D joint locations of 25 joints detected by Kinect-V2 depth sensors.
Each video has no more than 2 subjects. 
The original paper~\cite{shahroudy_ntu_2016} of the dataset recommends two benchmarks: 
(1) Cross-subject (CS): the dataset in this benchmark is divided according to the subject. 
(2) Cross-view (CV): the dataset in this benchmark is divided according to the camera. 
We follow this convention and report the top-1 accuracy on both benchmarks.

\textbf{NTU-120~\cite{liu_ntu_2019}} is an extension of NTU-60, which is larger and more challenging. 
It contains 113,945 action clips in 120 action classes. 
The clips are performed by 106 volunteers in 32 camera setups. 
It recommends two benchmarks: cross-subject (CS) and cross-setup (CE). 
Cross-subject is the same with NTU-60.
Cross-setup means using samples with odd setup IDs for training and others for testing. 

\textbf{SHREC~\cite{de_smedt_shrec17_2017}} is a challenging dataset for skeleton-based gesture recognition. 
For skeleton-based recognition tasks, modeling gestures are similar with modeling human bodies. 
Thus, we choose this dataset to show the generalizability of our model for different skeleton format.
It contains 2800 gesture sequences performed by 28 subjects in two ways: using one finger to perform gestures or using the whole hand to perform gestures. 
It provides the 3D coordinates of 22 hand joints captured by Intel-Real-Sense depth camera.
It recommends two benchmarks: recognizing 14 rough gesture categories (14-cls) and recognizing 28 fine-grained gesture categories (28-cls). 

\subsection{Ablation Study}
We analyze the individual components and their configurations in the final architecture. Unless stated, the network architectures except for the modules being studied and training schemes are the same for all experiments. 

\paragraph{SEM-Net}
For SEM-Net, we first compare the proposed GR-CFSA with the other kinds of attention blocks. All kinds of attention blocks have the same number of parameters. 
As shown in Table~\ref{tab:attention}, ``unified" denotes the directed implementation of the self-attention for skeleton data, i.e., Eq.~\ref{eq:attentionU} and Eq.~\ref{eq:weightU}. 
It has larger GFLOPs but lower performance. 
It may because that it models correlations cross too much elements, which is hard to learn. 
``single-frame" denotes considering only the intra-frame correlations, i.e., Eq.~\ref{eq:attention} and Eq.~\ref{eq:weight}.  
It needs only one fifth of the computational cost compared with ``unified", but it achieves higher performance. 
Compared with ``single-frame", our propose ``cross-frame" method, i.e., Eq.~\ref{eq:newweight}, achieves better performance with similar GFLOPs, which confirms the necessity of modeling inter-frame correlations. 
Finally, by adding the proposed global regularization (GR), i.e., Eq.~\ref{eq:gloreg}, the performance further improves, which shows the importance of the GR. 

\begin{table}[!htp]
    \centering
    \caption{Comparison for different attention blocks on the CS benchmark of NTU-60. }
    \label{tab:attention}
    \vspace{0.2cm} 
    \begin{tabular}{lccc}
    \hline\hline
    Settings & CS (\%) & CV (\%) & GFLOPs\\
    \hline
    unified             & 85.3 & 91.4 & 39.7 \\
    single-frame        & 88.4 & 94.3 & 7.5  \\
    cross-frame         & 88.9 & 94.9 & 7.4  \\
    \textbf{cross-frame+GR}      & \textbf{90.5} & \textbf{95.6} & \textbf{7.4} \\
    \hline\hline
    \end{tabular}
\end{table}

Then, we compare the different settings for feed-forward layer. 
As shown in Table~\ref{tab:ff}, enlarging the temporal kernel size (``k=3" to ``k=9") improves the performance. It is because it brings larger temporal receptive filed of the model. 
However, it also brings more GFLOPs and more parameters. 
By expanding it into the multi-scale manner (i.e., ``k=(3,5,7,9)"), it has the same temporal receptive filed with ``k=9", but the temporal window is more diverse to adapt to different samples.
Thus, the performance is further improved, while the GFLOPs and parameters drops due to the splitting of the channels. 
Finally, by changing the temporal dilation instead of the temporal kernel size, ``d=(1,3,5,7)" has the same temporal receptive filed with ``k=(3,5,7,9)", but it achieves the best performance and needs the least GFLOPs and parameters. 
\begin{table}[!htp]
    \centering
    \caption{Comparison for different settings for feed-forward layer on the CS benchmark of NTU-60. ``k" denotes the temporal kernel size, which is 3 by default. ``d" denotes the temporal dilation, which is 1 by default. }
    \label{tab:ff}
    \vspace{0.2cm} 
    \begin{tabular}{lcccc}
    \hline\hline
    Settings & CS (\%) & CV (\%) & GFLOPs & Params \\
    \hline
    k=3         & 88.6 & 94.2 & 9.2 & 6.4M \\
    k=9         & 89.3 & 95.1 & 13.9 & 10.5M \\
    k=(3,5,7,9) & 90.3 & 95.5 & 8.0 & 5.3M \\
    \textbf{d=(1,3,5,7)} & \textbf{90.5} & \textbf{95.6} & \textbf{7.4} & \textbf{4.8M} \\
    \hline\hline
    \end{tabular}
\end{table}

Table~\ref{tab:bone} shows the final accuracy of the SEM-Net on NTU-60.
Similar with the previous works, we fuse the joint features and the bone features, which brings about 1\% improvement. 
\begin{table}[!htp]
    \centering
    \caption{Recognition accuracy (\%) of SEM-Net on NTU-60. }
    \label{tab:bone}
    \vspace{0.2cm} 
    \renewcommand\tabcolsep{7.0pt} 
    \begin{tabular}{lcc}
    \hline\hline
    Settings & CS (\%) & CV (\%) \\
    \hline
    SEM-joint       & 90.5  & 95.6 \\
    SEM-bone        & 90.2  & 94.5 \\
    \textbf{SEM-fuse}        & \textbf{91.8}  & \textbf{96.3} \\
    \hline\hline
    \end{tabular}
\end{table}

\paragraph{SPA-Net.}
For SPA-Net, as shown in Table~\ref{tab:spa}, 
SPA-Net can achieve good classification performance, which verified the feasibility of the spatial perspective for skeleton data. 
Besides, ``SPA-4D" achieves better performance than ``SPA-3+1D". 
It may because the 4D convolution can directly model the spatial-temporal joint dependencies from low-levels to high-levels, while the 3+1D convolution will cause the semantic gap between the spatial dimensions and the temporal dimension. 

\begin{table}[!htp]
    \centering
    \caption{Recognition accuracy (\%) of SPA-Net on NTU-60. }
    \label{tab:spa}
    \vspace{0.2cm} 
    \renewcommand\tabcolsep{7.0pt} 
    \begin{tabular}{lcc}
    \hline\hline
    Settings & CS (\%) & CV (\%) \\
    \hline
    SPA-3+1D        & 89.2  & 94.9 \\
    \textbf{SPA-4D}          & \textbf{89.6}  & \textbf{95.5} \\
    \hline\hline
    \end{tabular}
\end{table}

The most important two factors that affect the performance of SPA-Net is the space size (s) and the kernel size (k). 
Space size affect the density of the joints in the space. Smaller size makes many neighboring joints falling into the same position, which causes the loss of information. 
Larger size makes the distance between the joints too far, which brings difficulties
for relation modeling with the fix-size convolutional kernel. 
Based on the experiments on Tab.~\ref{tab:sk}, ``s=64" is a good choice. 
Kernel size affect the size of the neighborhood in each convolutional step. Because the joints are sparse, larger kernel size can help covering more points so that
capture more information in one convolutional step. 
However, too large kernel size will only lead to the waste of parameters. 
According to the experiments, ``k=5" is the best choice. 
\begin{table}[!htp]
    \centering
    \caption{Comparison of different settings for SPA-Net on NTU-60. ``k" denotes the kernel size, and ``s" denotes the space size. }
    \label{tab:sk}
    \vspace{0.2cm} 
    \renewcommand\tabcolsep{7.0pt} 
    \begin{tabular}{lcc}
    \hline\hline
    Settings & CS (\%) & CV (\%) \\
    \hline
    s=32, k=5 & 88.5 & 94.7 \\
    \textbf{s=64, k=5} & \textbf{89.6} & \textbf{95.5} \\
    s=128, k=5 & 89.5 & 95.3 \\
    s=64, k=3 & 88.6 & 94.5 \\
    s=64, k=6 & 89.4 & 95.3 \\
    \hline\hline
    \end{tabular}
\end{table}

\paragraph{Fusion.}
As shown in Table~\ref{tab:fusion}, fusion of the two perspectives (``SEM+SPA") obtains notable improvements (0.8\%/0.8\% compared with SEM-Net and 3.0\%/1.6\% compared with SPA-Net). 
It demonstrates the complementarity of the two perspectives and the necessarily of the fusion. 
Note that SEM already fuses the bone information, and 
By using only the joint information, ``SEM-joint+SPA" brings more improvements (1.5\%/1.3\%) compared with ``SEM-joint", and also achieves better performance than the joint-bone fusion manner, i.e., ``SEM-Net". 

\begin{table}[!htp]
    \centering
    \caption{Fusion of the SEM-Net and SPA-Net on NTU-60. }
    \label{tab:fusion}
    \vspace{0.2cm} 
    \renewcommand\tabcolsep{7.0pt} 
    \begin{tabular}{lcc}
    \hline\hline
    Settings & CS (\%) & CV (\%) \\
    \hline
    SPA-Net             & 89.6  & 95.5 \\
    SEM-Net             & 91.8  & 96.3 \\
    \textbf{SEM+SPA }        & \textbf{92.6}  & \textbf{97.1} \\
    SEM-joint       & 90.5  & 95.6 \\
    SEM-joint+SPA   & 92.0  & 96.9 \\
    \hline\hline
    \end{tabular}
\end{table}

\subsection{Visualization}
To better show the difference between the SEM-Net and the SPA-Net, 
we plot their learned features for the sample ``sit down" as shown in Fig.~\ref{fig:fea}. 
More visualizations are provided in the supplement material. 
It shows the SEM-Net focuses more on mining important relations between different human parts, while the SPA-Net focuses more on mining discriminative features in the local areas. 
In detail, for SEM-Net, we draw the top-50 attention weights as the orange lines in (b). 
It shows SEM-Net takes stronger attentions for correlations of hands and spline base, which is the characteristics of the action being identified. 
For SPA-Net, we draw the activations in the 3D space in (c), where the circle size represents the magnitude of the activation. 
It shows SPA-Net also activates more on the discriminative areas of action ``sit down", i.e., the areas containing the hands and the areas containing the spline base, especially when the person is sitting down (third column).
These visualizations confirmed our motivation and also showed the effectiveness of our methods. 

Besides, we also plot the per-class accuracies of SEM-Net and SPA-Net on NTU-60 (CS) and SHREC (14-cls). Due to the space limitation, the figure of NTU-60 are provided in the supplement material. 
It shows the two models preforms different for different categories, which shows the complementarity of the two perspectives. 

\begin{figure}[!htb]
    \centering
    \includegraphics[width=1\linewidth]{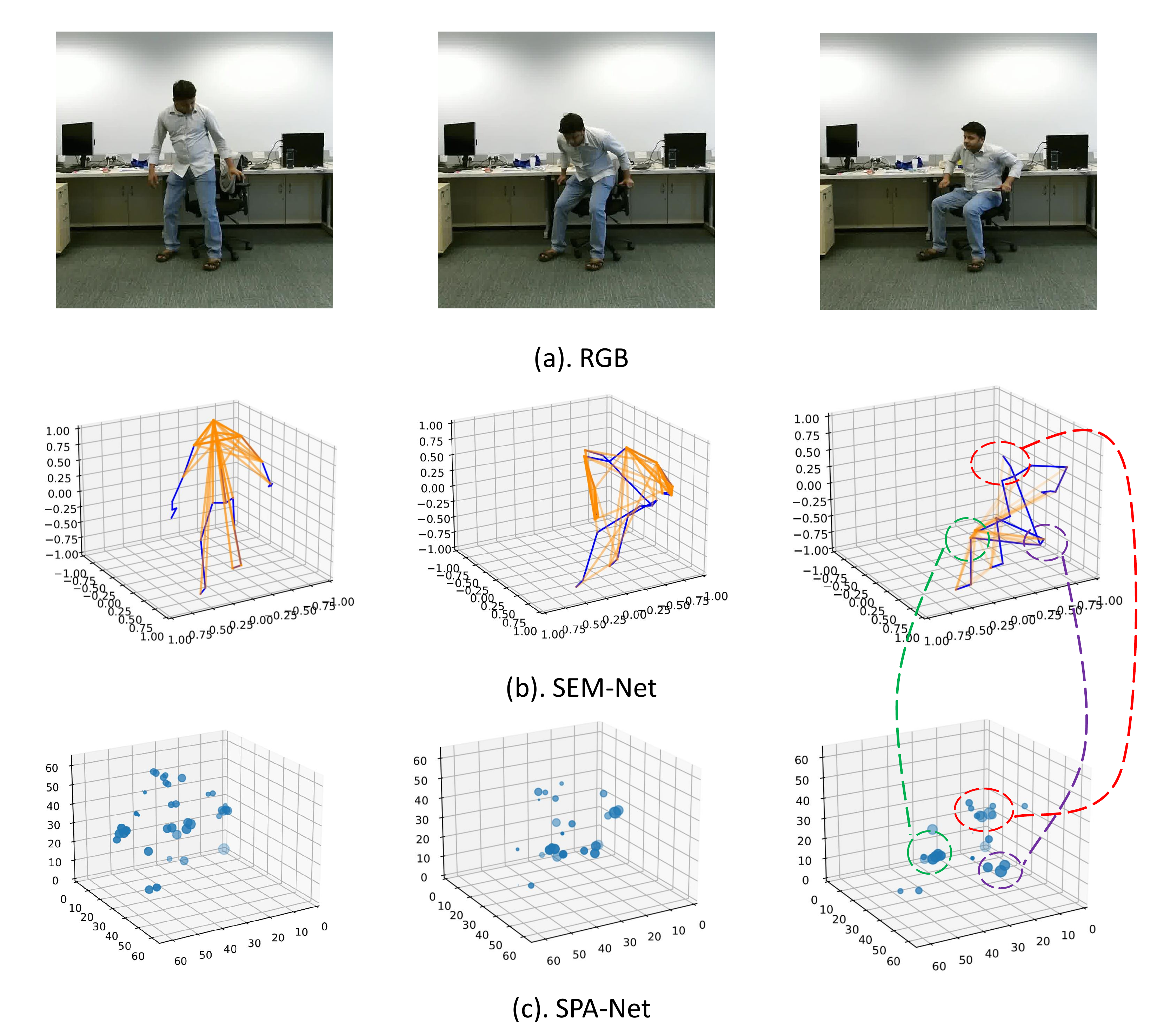}
    \caption{(a) shows the original RGB images of the sample ``sit down". (b) shows the corresponding skeletons (blue lines) and the learned attentions weights (orange lines) by SEM-Net. 
    (c) shows the activations of the SPA-Net, where the circle size represents the magnitude of the activation. 
    }
    \label{fig:fea}
\end{figure}

\begin{figure}
    \centering
    \includegraphics[width=\linewidth]{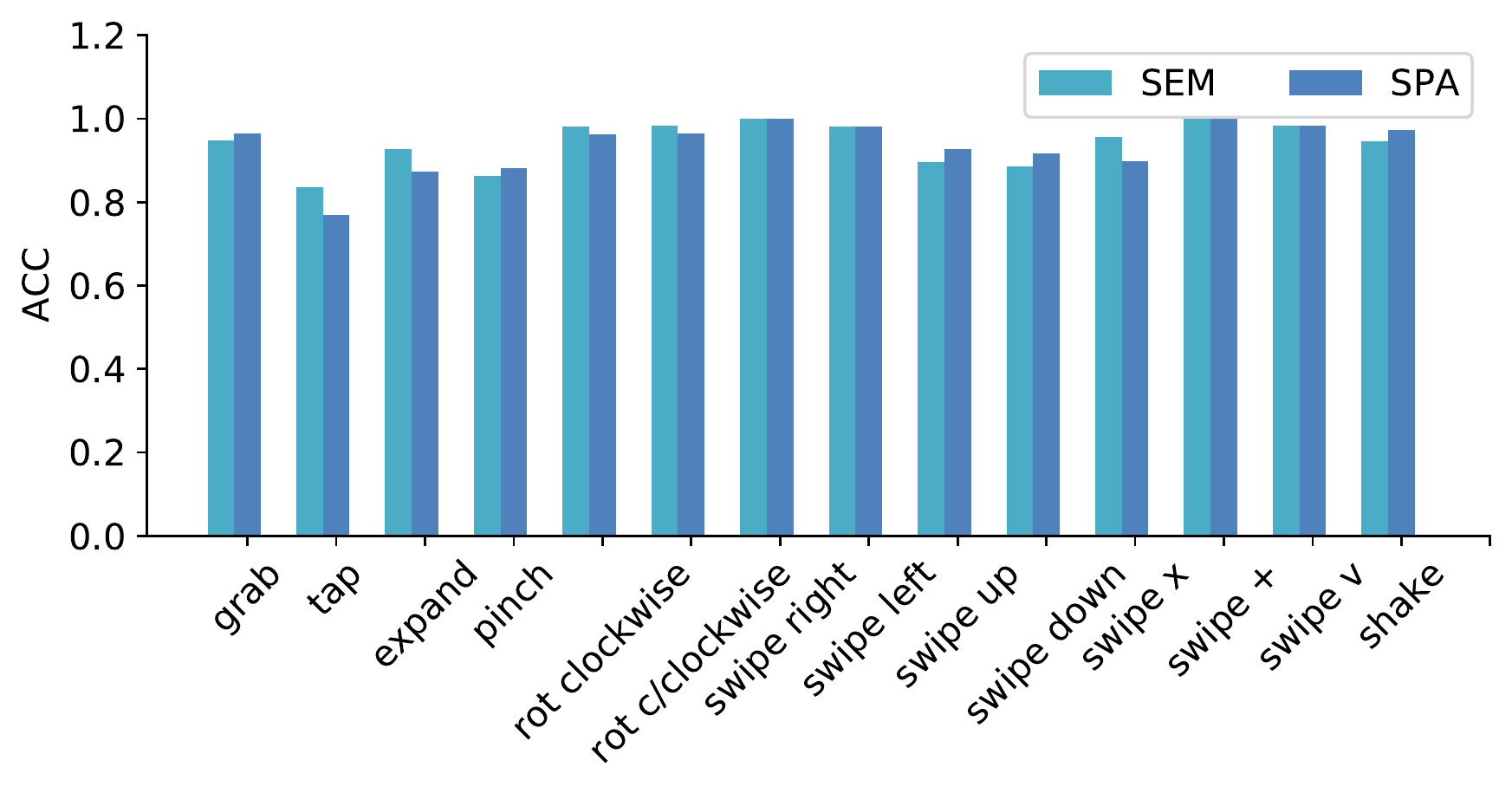}
    \caption{Per-class accuracies of SEM-Net and SPA-Net on SHREC. 
    }
    \label{fig:shrec}
\end{figure}

\begin{table}[!htp]
    \centering
    \caption{Comparison with the state-of-the-arts on NTU. }
    \label{tab:ntu}
    \vspace{0.2cm} 
    \renewcommand\tabcolsep{7.0pt} 
    \begin{tabular}{lccc}
    \hline\hline
    Methods & Year & CS (\%)  & CV (\%)  \\
    \hline
    2s-AGCN~\cite{shi_two-stream_2019} & 2019  & 88.5 & 95.1 \\
    DGNN~\cite{shi_skeleton-based_2019}  & 2019 & 89.9 & 96.1 \\
    DCM+SAM~\cite{xiao_self-attention_2019} & 2019 & 86.2 & 92.2 \\
    \hline
    CA-GCN~\cite{zhang_context_2020} & 2020 & 86.5 & 94.1 \\
    TS-SAN~\cite{cho_self-attention_2020} & 2020 & 87.2 & 92.7 \\
    SGN~\cite{zhang_semantics-guided_2020} & 2020 & 89.0 & 94.5 \\
    NAS~\cite{peng_learning_2020}  & 2020 & 89.4 & 95.7 \\
    MS-G3D~\cite{liu_disentangling_2020} & 2020 & 91.5 & 96.2 \\ 
    \hline
    \textbf{SEM+SPA (ours)}     & - & \textbf{92.6} & \textbf{97.1} \\ 
    \hline\hline
    \end{tabular}
\end{table}

\subsection{Comparison against the State-of-the-Art}
We compare our full model to the state-of-the-arts in Tab.~\ref{tab:ntu}, Tab.~\ref{tab:NTU120} and Tab.~\ref{tab:shrec}. 
On all three datasets, our method outperforms the state-of-the-art methods with a significant margin. 
Note that even not using the SPA-Net, our method still performs better than previous works.  

\begin{table}[!htb]
    \centering
    \caption{Comparison with the state-of-the-arts on NTU-120.}
    \label{tab:NTU120}
    \vspace{0.2cm} 
    \renewcommand\tabcolsep{3.0pt} 
    \begin{tabular}{l cc cc}
        \hline\hline
        Methods & Year & CS (\%)  & CE (\%)  \\
        \hline
        SkeMotion~\cite{caetano_skelemotion_2019} & 2019 & 67.7 & 66.9 \\
        AGCN~\cite{shi_two-stream_2019} & 2019 & 82.9 & 84.9\\
        SGN~\cite{zhang_semantics-guided_2020} & 2020   & 79.2  & 81.5 \\
        MS-G3D~\cite{liu_disentangling_2020} & 2020 & 86.9 &  88.4  \\
        \hline
        \textbf{SEM (ours)}        & - & \textbf{87.5} & \textbf{89.4} \\ 
        \textbf{SPA (ours)}         & - & \textbf{83.8} & \textbf{85.2} \\ 
        \textbf{SEM+SPA (ours)}     & - & \textbf{88.7} & \textbf{90.3} \\ 
        \hline\hline
    \end{tabular}
\end{table}

\begin{table}[!htp]
    \centering
    \caption{Comparison with the state-of-the-arts on SHREC. }
    \label{tab:shrec}
    \vspace{0.2cm} 
    \renewcommand\tabcolsep{5.0pt} 
    \begin{tabular}{lccc}
    \hline\hline
    Methods & Year & 14-cls (\%) & 28-cls (\%) \\
    \hline
    ST-GCN~\cite{yan_spatial_2018} & 2018 &  92.7 & 87.7  \\
    STA-Res-TCN~\cite{hou_spatial-temporal_2018} & 2018 & 93.6 & 90.7  \\
    ST-TS-HGR-NET~\cite{nguyen_neural_2019} & 2019 &  94.3 & 89.4  \\
    DG-STA.~\cite{chen_construct_2019} & 2019 & 94.4 & 90.7  \\
    HPEV~\cite{liu_decoupled_2020}  & 2020 & 94.9 & 92.3  \\
    \hline
    \textbf{SEM (ours)  }      & - & \textbf{95.0} & \textbf{93.9} \\ 
    \textbf{SPA (ours) }        & - & \textbf{93.6} & \textbf{91.9} \\ 
    \textbf{SEM+SPA (ours)}     & - & \textbf{96.2} & \textbf{94.4} \\ 
    \hline\hline
    \end{tabular}
\end{table}


\section{Conclusion}
In this work, we propose a novel framework for skeleton-based action recognition, where the skeletons are modeled from both the semantic perspective and the spatial perspective. 
Besides, we customized two networks to adapt the two perspectives for skeleton modeling. 
The proposed method is verified on three challenging datasets for skeleton-based human action/gesture recognition, where our method achieves the state-of-the-art performance on all of them. 
Future works can be focused on how to fuse the two perspectives in earlier stage to make the fusion more adequate. Besides, it is also worth to explore whether it is possible to transform the two perspectives into a unified form.  

{\small
\bibliographystyle{ieee_fullname}
\bibliography{references}
}

\end{document}